%% file: root.tex
\documentclass[letterpaper, 10 pt, conference]{ieeeconf}

\IEEEoverridecommandlockouts       
\makeatletter
\let\NAT@parse\undefined
\makeatother

\usepackage{hyperref}

\usepackage{subcaption}
\usepackage{amsmath} 
\usepackage{amssymb} 
\usepackage{graphicx, xcolor}
\usepackage{bm}
\usepackage{balance}
\usepackage{cite}
\usepackage{tabularx}
\usepackage{siunitx}

\usepackage[utf8]{inputenc}
\usepackage{pgfplots}
\DeclareUnicodeCharacter{2212}{−}
\usepgfplotslibrary{groupplots,dateplot}
\usetikzlibrary{patterns,shapes.arrows}
\pgfplotsset{compat=newest}

\hypersetup{
  colorlinks    = true, 
  urlcolor      = blue, 
  linkcolor     = blue, 
  citecolor     = red   
}
\newcommand{\textref}[2]{\hyperref[#1]{#2~\ref*{#1}}}

\title{\LARGE \bf
ProDapt: Proprioceptive Adaptation using Long-term Memory Diffusion}

\author{Federico Pizarro Bejarano$^{*1,2}$, Bryson Jones$^{2}$, Daniel Pastor Moreno$^{2}$,
Joseph Bowkett$^{2}$, \\ Paul G. Backes$^{2}$, Angela P. Schoellig$^{1,3}$
\thanks{$^*$ Corresponding author:  federico.pizarrobejarano@mail.utoronto.ca}%
\thanks{$^1$ University of Toronto and Vector Institute for Artificial Intelligence}%
\thanks{$^2$ Jet Propulsion Laboratory, California Institute of Technology}%
\thanks{$^3$ Technical University of Munich~(TUM) and Munich Institute for Robotics and Machine Intelligence~(MIRMI)}
}

\begin{document}

\maketitle
\thispagestyle{empty}
\pagestyle{empty}

\begin{abstract}
Diffusion models have revolutionized imitation learning, allowing robots to replicate complex behaviours. However, diffusion often relies on cameras and other exteroceptive sensors to observe the environment and lacks long-term memory. In space, military, and underwater applications, robots must be highly robust to failures in exteroceptive sensors, operating using only proprioceptive information. In this paper, we propose ProDapt, a method of incorporating long-term memory of previous contacts between the robot and the environment in the diffusion process, allowing it to complete tasks using only proprioceptive data. This is achieved by identifying ``keypoints'', essential past observations maintained as inputs to the policy. We test our approach using a UR10e robotic arm in both simulation and real experiments and demonstrate the necessity of this long-term memory for task completion. 
\end{abstract}

\section{Introduction}
Imitation learning aims to learn a control policy directly from observing others perform a task. However, imitation learning faces many challenges, including high-dimensional observation and action spaces, multi-modality, and strong temporal correlation. These challenges have limited the performance of imitation learning on complex and precise tasks. 

Diffusion has been successfully applied to imitation learning~\cite{diffusion_policy, flexible_diffusion}. Diffusion~\cite{ddpm} is a machine learning framework that generates samples of an unknown distribution, trained using samples from the distribution. The generative process is modelled as an iterative denoising process. Diffusion's success is due to its high expressivity, ability to tackle high-dimensional spaces, and stable training. 

\begin{figure}[t]
  \centering
  \includegraphics[width=1.0\linewidth]{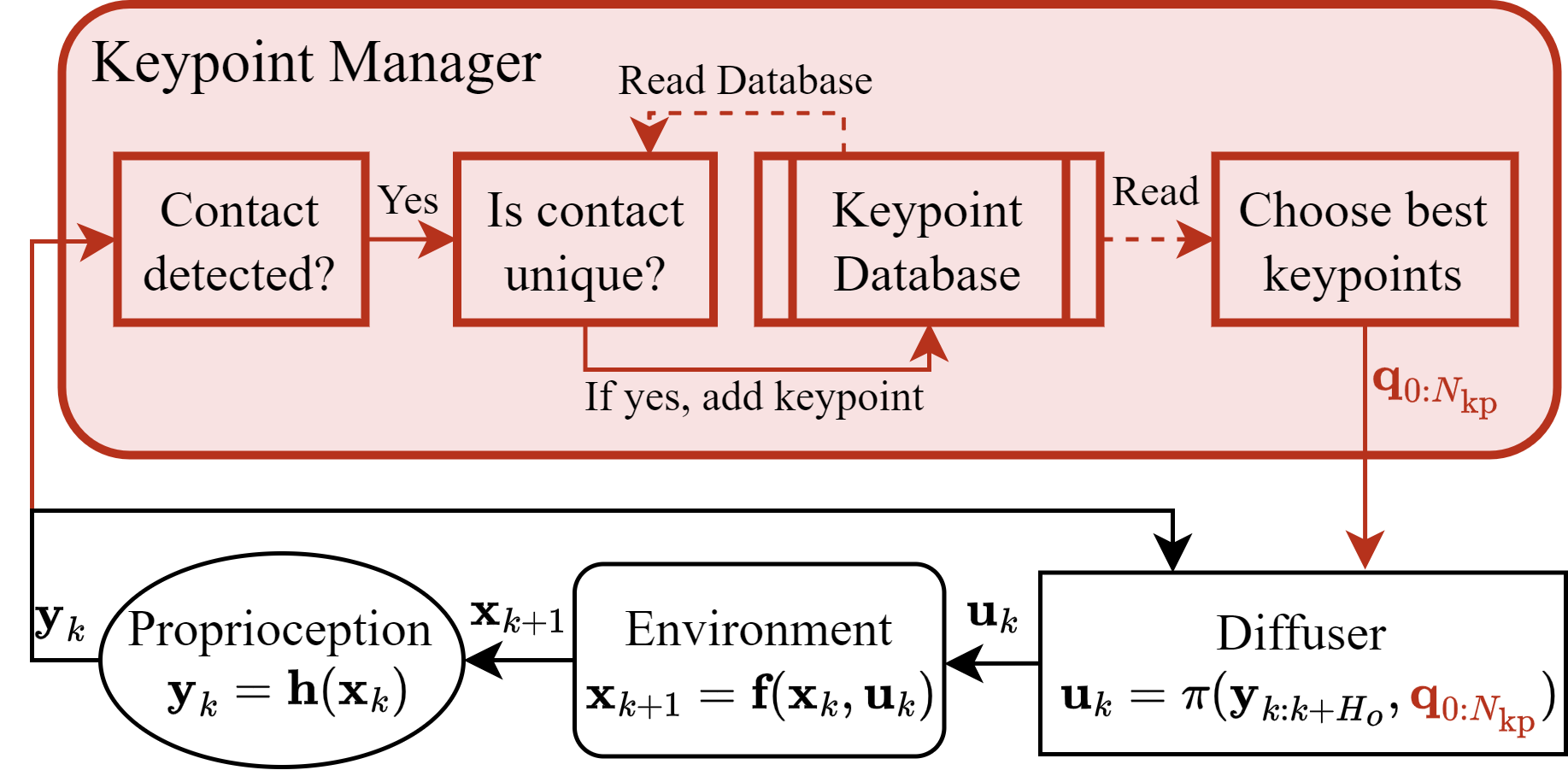}\\
  \caption{Our proposed approach with the novel components in red. The proprioceptive sensor measurements $\textbf{y}_k$ are provided to the keypoint manager, which determines if the robot has contacted an obstacle and whether that contact is unique. If so, that contact is saved as a keypoint. During inference, the diffusion model is conditioned on the most useful keypoints and previous observations and outputs the next actions.}
  \label{fig:summary_diagram}
  \vspace{-4mm}
\end{figure} 

Diffusion has been applied to robotics by generating action sequences given past observations~\cite{flexible_diffusion, diffusion_policy}. It has been applied to manipulation~\cite{diffusion_policy}, soft robots~\cite{soft-robot}, and locomotion~\cite{locomotion}. However, existing approaches rely on visual information or task-specific state estimation~\cite{3d_diff}. Operating with only proprioceptive sensors is necessary for tasks with adverse conditions which may compromise exteroceptive sensors, such as space applications, or situations in which exteroceptive sensors are unable to perceive the environment effectively, such as space and underwater applications where mud and dust often occlude cameras~\cite{eels1, eels2}. 

\textit{Contributions}: We propose a method of incorporating long-term memory into learning-based controllers. Past observations essential for reasoning about the present task are maintained as ``keypoints``and are applied to the controller as inputs. In the case of proprioception, previous contacts with obstacles are recorded and included in the conditioning of a diffusion model to allow for operation without exteroceptive data (see \textref{fig:summary_diagram}{Fig.}). Our code and results are available at \hphantom{....} \url{https://tinyurl.com/prodapt-code}, and a video of our experiments is available at \url{https://tinyurl.com/prodapt-video}.

\section{Related Work}

\subsection{Diffusion}
Diffusion models complex systems and generates high-quality data samples by transforming a simple distribution (usually the standard Gaussian distribution) into a target distribution (such as a distribution of real-world images) through a series of gradual steps~\cite{ddpm}. Diffusion boasts stable training and high expressivity, multi-modality, and interpretability. 

During training, the forward diffusion process progressively adds noise to real data. This process transforms a real data sample into a sample from a simple, known distribution:
\begin{align}
    \textbf{x}^0 \rightarrow \textbf{x}^1 \rightarrow \textbf{x}^2 \rightarrow \dots \rightarrow \textbf{x}^{N_{\text{diff}}},
\end{align}
where $\textbf{x}^0$ is the original data sample, $\textbf{x}^i$ is $\textbf{x}^0$ corrupted by noise $i$ times, $N_{\text{diff}}$ is the number of diffusion steps, and $\textbf{x}^{N_{\text{diff}}}$ is a sample from the standard Gaussian distribution.

The reverse process gradually denoises the noisy samples to recover the original data. A neural network parameterizes the denoising operation:
\begin{align}
    \textbf{x}^{N_{\text{diff}}} \rightarrow \hat{\textbf{x}}^{N-1} \rightarrow \dots \rightarrow \hat{\textbf{x}}^1 \rightarrow \hat{\textbf{x}}^0,
\end{align}
where $\hat{\textbf{x}}^{N_{\text{diff}}-i}$ is the Gaussian sample $\textbf{x}^{N_{\text{diff}}}$ after $i$ denoising steps, and $\hat{\textbf{x}}^0$ is the recovered data sample.

Diffusion has demonstrated state-of-the-art image~\cite{ddpm} and video generation~\cite{diffusion_video}. In~\cite{flexible_diffusion}, diffusion was extended to robot control by viewing a matrix of past observations and actions as a distribution, generating matrices of future observations and actions. This approach can incorporate reward-shaping, varying-length trajectories, and constraints. In~\cite{diffusion_policy}, the performance of diffusion models was improved by conditioning on observations rather than predicting actions and observations jointly, as well as a receding-horizon approach for real-time control. Many other extensions of diffusion for control have been published, including using 3D point clouds~\cite{3d_diff} and semantic understanding~\cite{dall-e-bot}.

\subsection{Proprioceptive Planning}
A standard approach to mapping areas with robots before the widespread use of cameras or LiDAR was wall-following~\cite{tactile-nav}, which is still an active area of research~\cite{wall-follower}. Wall-following has been extended to blind road following~\cite{spider-road-follower} and wire following~\cite{wire-following}. However, wall-following is a suboptimal approach to blind navigation~\cite{maps-emerge}. Extreme fault tolerance~\cite{eels1, eels2, cio} ensures robot safety and operability when exteroceptive sensors fail. These approaches rely on redundant proprioception-only navigation loops. However, there is relatively little research in full motion planning for blind robots, which is the target of this paper.


In this work, we aim towards navigation of cluttered environments relying on proprioceptive sensors. In~\cite{tactile-nav}, a tactile navigation algorithm leverages pre-designed action macros, Gaussian mixture models to classify features, and explicit mapping to navigate a simulated indoor space. In comparison,~\cite{maps-emerge} demonstrated that blind agents using a long short-term memory (LSTM) recurrent neural network controller with no explicit mapping can outperform other blind navigation agents, performing similarly to agents with exteroception. Our approach is conceptually similar to~\cite{maps-emerge} using only a diffusion model to complete the entire navigation stack, allowing for generalizable task completion. This is done by leveraging the expressivity of diffusion to learn the appropriate reactions to past and present collisions during task execution and maintaining a memory of past contacts.

\subsection{Long-Term Memory in Machine Learning}
Several machine learning approaches incorporate long-term memory. Recurrent architectures like long short-term memory~\cite{lstm} and gated recurrent units (GRU)~\cite{gru} were foundational but struggle with long-range dependencies due to vanishing gradients~\cite{transformers, lstm_comp}. Transformers~\cite{transformers}, particularly architectures like RetNet~\cite{retnet} and RWKV~\cite{rwkv}, improve memory efficiency by introducing structured recurrence within the self-attention framework, allowing them to handle long-term dependencies effectively. More recent approaches, such as Mamba~\cite{mamba}, leverage structured state-space models (SSMs)~\cite{ssm} to provide implicit long-term memory through continuous state evolution, maintaining computational efficiency while processing long sequences. Existing generative models, including diffusion models, do not retain long-term memory.

\section{Problem Formulation}
We consider a discrete, time-invariant system given by
\begin{align} \label{eq:true-system}
    \textbf{x}_{k+1} = \textbf{f}(\textbf{x}_{k}, \textbf{u}_k),
\end{align}
where $\textbf{x}_k \in \mathbb{X}$ represents the system state at time step $k$, $\textbf{u}_k \in \mathbb{U}$ denotes the control input, and $\textbf{f}$ encapsulates the system dynamics and the environment.

We do not assume complete knowledge of the environment's state. Instead, we consider an observation model where at state $\textbf{x}_k$ we get observation $\textbf{y}_k$ given by
\begin{align}
    \textbf{y}_{k} = \textbf{h}(\textbf{x}_{k}),
\end{align}
where $\textbf{h}$ is represents the sensors. This paper considers only proprioceptive sensing, so reconstructing the environment from the latest observation $\textbf{y}_k$ is not possible. 

The diffusion controllers considered in this paper are based upon the architecture proposed in~\cite{diffusion_policy}, which uses receding control. Consider three horizons: the prediction horizon $H_p$, the observation horizon $H_o$, and the action horizon $H_a$. At a given time step $k$, the diffusion controller takes in $H_o$ previous observations and outputs the next $H_p$ actions:
\begin{align} \label{eq:diff_eq}
    \textbf{u}_{k:k+H_p-1} = \pi(\textbf{y}_{k-H_o+1:k})
\end{align}
where $\pi$ is the diffusion controller, and $\textbf{p}_{i:j} = [\textbf{p}_i, \textbf{p}_{i+1}, ..., \textbf{p}_j]$ for any variable $\textbf{p}$. Of the $H_p$ actions generated by the diffusion model, only $H_a$ actions are applied to the system before the next generation step. We recover the standard control approach if $H_o$ and $H_a$ are set to 1. This approach to receding horizon control allows for a tunable number of past observations $H_o$ passed to the diffusion model, trades off reactiveness with computation using $H_a$, and provides improved multi-modality and robustness to idle actions using $H_p$ and $H_a$~\cite{diffusion_policy}.

\section{Method} \label{sec:method}

\subsection{Motivating Example}
The Artemis program is an international project dedicated to establishing a permanent human base on the Moon, led by NASA~\cite{artemis}. The lunar base will provide a jumping-off point for further solar system exploration, including eventual human operations on Mars. For the Artemis project to succeed, it is vital to develop robots to assist in various tasks, notably construction and maintenance of the lunar base~\cite{artemis}. NASA has announced that general-purpose robotic manipulation is a critical technology that must be developed for the project to succeed ~\cite{shortfall}. 

Autonomous construction on the lunar surface suffers from several significant problems that may reduce the reliability of cameras and other exteroceptive sensors. Firstly, space technology is expected to be highly robust to failure, up to and including the complete failure of exteroceptive sensors~\cite{eels1, eels2}. Additionally, the lack of a lunar atmosphere and low gravity on the Moon result in dust clouds when the surface is disturbed, occluding cameras~\cite{eels1, eels2}. Despite damage or occlusion of exteroceptive sensors, the robot must be capable of reliable and safe performance. 

For this application and other similar applications, such as the ocean where mud and sand can occlude the camera, our motivating example is safely moving an object or end-effector from a starting point to a goal area despite unknown obstacles, without exteroceptive sensors.

\subsection{Keypoints}
Consider moving from a starting point to a goal point with a single obstacle between the two points without exteroceptive sensors. A naive approach that only takes in the current observation (i.e., $H_o=1$)  may fail at this task; it will collide with the obstacle and move backwards, but immediately ``forget'' it collided with an obstacle and move forward, colliding again. This oscillation will continue until the robot, due to chance or drift, passes the obstacle. Using a longer observation horizon $H_o$ may improve performance as the robot will ``remember'' the collision for $H_o$ time steps. However, if it cannot pass the obstacle in $H_o$ time steps, it will forget the obstacle, potentially leading the robot to be similarly trapped. Additionally, increasing $H_o$ will incur a higher computational cost, which is especially undesirable in space applications with limited computational resources. 

Long-term memory of the environment is necessary to navigate around obstacles. A classical approach is to create a map of the environment using the contact points. However, mapping from sparse contact points of obstacles is complicated and unnecessary. Instead, we leverage the expressivity of diffusion models to learn to react to past and present contacts. A curated list of contact points can be maintained and provided to the diffusion model as a ``memory'' of contacts. We will denote these recorded contacts as ``keypoints''. Recording relevant and informative past observations and applying them as inputs to a learning-based controller can be expanded beyond diffusion and proprioception, such as recording important waypoints reached in a trajectory. In this way, we present a generalizable approach to incorporating long-term memory into learning-based controllers. 

A keypoint, the $i\text{-th}$ of which we denote as $\textbf{q}_i$, contains the relevant information for the task. In proprioceptive navigation, we record the contact position and the force and torque experienced due to the collision. What determines a contact depends on the sensors; on a robotic arm equipped with force-torque sensors, a contact is a point where those forces spike unexpectedly, while in an application without those sensors, a contact has occurred when the robot cannot move forward despite applied force. 

To reduce computation, we provide the diffusion model with the minimum necessary number of keypoints by retaining only sufficiently different keypoints. Additionally, keypoints that are likely uninformative in the current situation (e.g., ones that are very distant from the current position of the robot) may be excluded from the diffusion inference. The system which determines which states represent a contact, determines if the contact is sufficiently different from existing keypoints, saves the vital information into a keypoint, and provides only the necessary keypoints for each diffusion step is referred to as the ``keypoint manager''.

\subsection{Diffusion with Keypoints}
After a keypoint manager has been developed and tuned for the specific robotic application, it is necessary to incorporate the keypoints into the diffusion model. However, there are a varying number of keypoints at each time step. A maximum number of keypoints $N_{\text{kp}}$ is set, and this number of keypoints is passed to the diffusion model at each inference step, appending to the end of the list of past observations. If there are fewer informative keypoints than the maximum, the unset keypoints are passed as zeros (i.e., zero-padding~\cite{zero-padding}). We modify \textref{eq:diff_eq}{Eq.} to include up to $N_{\text{kp}}$ keypoints: 
\begin{align}
    \textbf{u}_{k:k+H_p-1} = \pi(\textbf{y}_{k-H_o+1:k}, \; \textbf{q}_{1:N_{\text{kp}}}).
\end{align}

Our diffusion models use a convolutional neural network~(CNN) U-Net diffusion architecture~\cite{ddpm}. However, when using a transformer-based diffusion architecture~\cite{diffusion_policy}, the unset keypoints can be replaced with padding tokens rather than zero-padding. 

\section{Experimental Results}
To demonstrate the importance of long-term memory of contacts, we ran experiments on a Universal Robots UR10e manipulator arm, both real and simulated in NVIDIA Isaac Sim~\cite{isaac} (see \textref{fig:sim_v_real}{Fig.}). The diffusion models are based upon the convolutional neural network U-Net diffusion architecture architecture proposed in~\cite{diffusion_policy}. We compare our keypoint approach with a standard memory-less diffusion model with observation horizons $H_o=3, 6, 20, 50$, which we denote as the baseline approaches. Our approach has $H_o=3$ and ten keypoints, as discussed below. The only differences between our approach and the baselines are the different observation horizons and the addition of keypoints. 

\begin{figure}[t]
    \centering
    \vspace{2mm}
    \begin{subfigure}{.49\linewidth}
        \includegraphics[width=\linewidth]{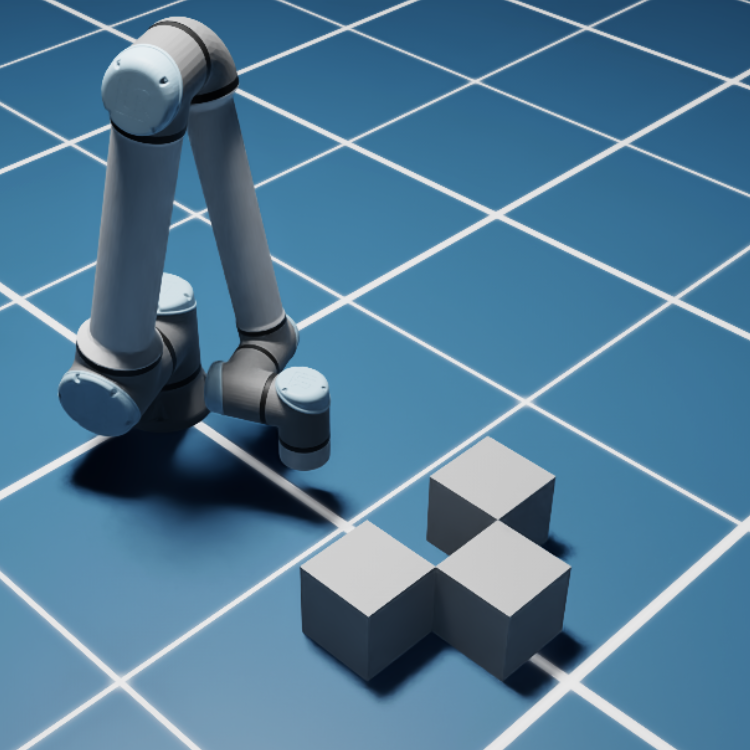}
        \caption{Simulation in Isaac Sim}
    \end{subfigure}\hfill
    \begin{subfigure}{.49\linewidth}
        \includegraphics[width=\linewidth]{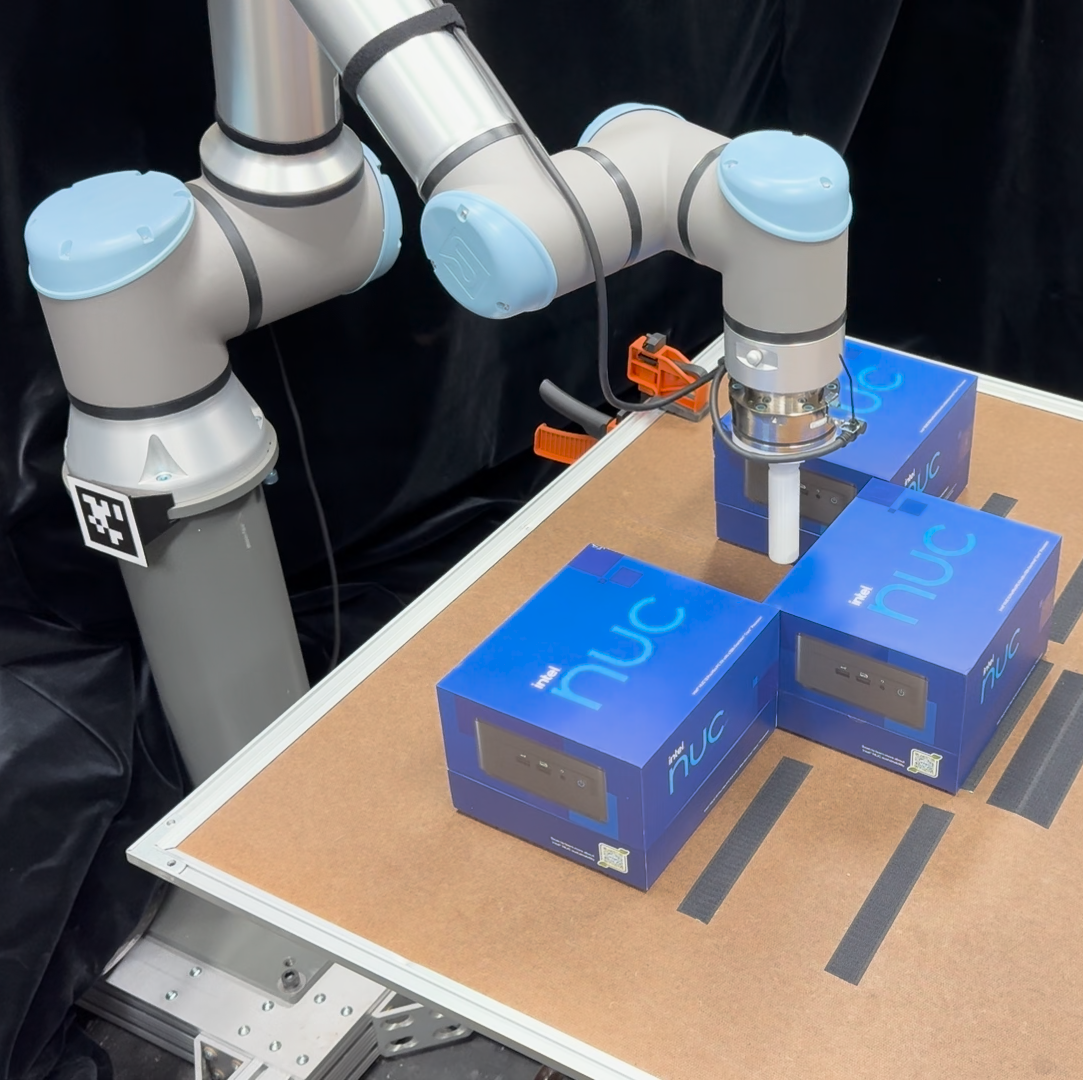}
        \caption{Real UR10e}
    \end{subfigure}
    \caption{Experimental setup on a simulated UR10e in Isaac Sim and a real UR10e. In both setups, the arm must move from a starting position to a goal position despite unknown obstacles and no exteroception.}
  \label{fig:sim_v_real}
  \vspace{-4mm}
\end{figure} 

The simulated and real experiments reflect our motivating example. The UR10e must move its end-effector from a starting point to a goal point (restricted to the $x-y$ plane for simplicity) despite random unknown obstacles. The task was designed to demonstrate the importance of long-term memory and is well-suited for diffusion as it is highly multi-model; for example, a robot can pass an obstacle on the right or left. Additionally, diffusion offers strong robustness to changes in the environment, which is essential for executing models trained with simulation data on real hardware. Finally, diffusion provides smooth trajectories due to generating trajectories of length $H_p$, which is important for smooth movements of the UR10e arm. 

The UR10e has a force-torque sensor at the end-effector's joint, and contact is flagged when the torque applied in the plane exceeds a threshold, $\tau_{\text{min}} = $ \SI{1}{\newton\meter}. Once a contact is detected, the contact is saved as a keypoint if it is at least $d_{\text{min}} = $ \SI{5}{\centi\meter} away from every recorded keypoint. Otherwise, the contact is only saved if the direction of the normal force of the contact is a sufficient angular distance $\theta_{\text{min}} = $ \SI{45}{\degree} away from the keypoints less than $d_{\text{min}}$ distance nearby, such as would be found in a corner. In our experiments, we set the maximum number of keypoints $N_{\text{kp}}$ to return to the diffusion model at ten and keep the last ten keypoints rather than maintaining a more extensive database and returning a curated list. Each keypoint comprises the $x-y$ position of the end-effector during the contact and the sine and cosine of the normal direction of the contact. 

The prediction horizon $H_p$ is set to 20, and the action horizon $H_a$ is set to 10. The solutions for the next diffusion inference step are warmstarted using the previous action trajectory according to~\cite{flexible_diffusion}. The diffusion model is conditioned on $H_o$ observations, each consisting of the $x-y$ position of the end-effector and the $x-y$ torque recorded by the end-effector's force torque sensor. Our diffusion approach is also conditioned upon the $N_{\text{kp}}=10$ last keypoints. The diffusion model then outputs $H_p$ actions, each of which is a $x-y$ goal position for the end-effector. We chose to use position control due to the recommendation in~\cite{diffusion_policy}. The total number of diffusion steps $N_{\text{diff}}$ is set to 45. 

The training of the diffusion controllers was done by collecting 165 teleoperation trajectories in the simulator of an operator completing the goal-seeking task with random obstacles, observing only the end-effector pose and force-torque readings. The obstacles were 15cm cubes with random orientations and uniformly distributed in the space $x \in$ [\SI{0.55}{\meter}, \SI{1.0}{\meter}], $y \in$ [-\SI{0.3}{\meter}, \SI{0.3}{\meter}]. The teleoperation was done using a 3D mouse with a control frequency of 10Hz, and the simulated and real experiments were also run at 10Hz. The end-effector always started at (\SI{0.4}{\meter}, \SI{0}{\meter}) and ended at the goal state, (\SI{1.2}{\meter}, \SI{0}{\meter}). For evaluation, the starting state was set to (\SI{0.7}{\meter}, \SI{0}{\meter}) so the arm could begin directly in front of the obstacles.

Each model was trained using an Nvidia A100 80GB GPU, partitioned to train two models in parallel. The training was done for 5000 epochs. The training time depended on $H_o$, with our approach and the $H_o=3$ approach taking approximately three hours and the $H_o=50$ approach taking approximately seven hours. Training hyperparameters can be found in our code repository. 

\begin{figure}[t]
    \centering
    \vspace{2mm}
    \begin{subfigure}{.49\linewidth}
        \includegraphics[width=\linewidth]{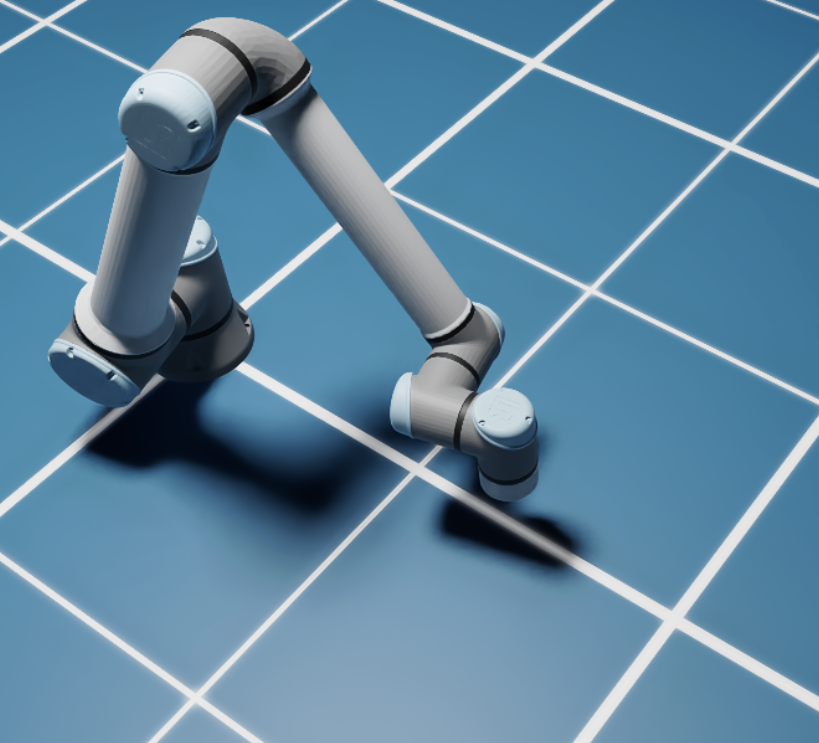}
        \caption{Clear}
    \end{subfigure}\hfill
    \begin{subfigure}{.49\linewidth}
        \includegraphics[width=\linewidth]{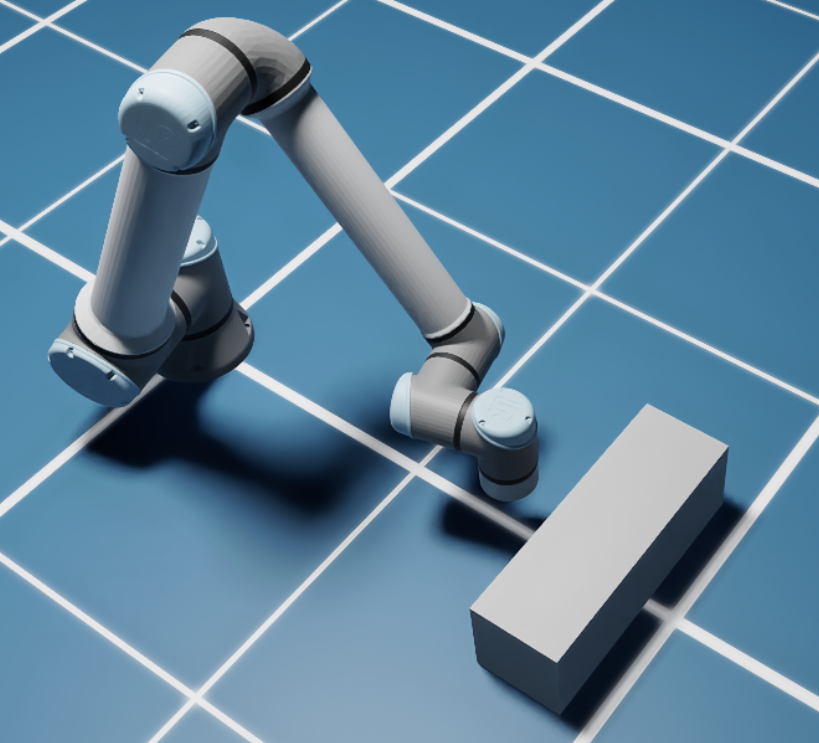}
        \caption{Wall}
    \end{subfigure}
    \begin{subfigure}{.49\linewidth}
        \includegraphics[width=\linewidth]{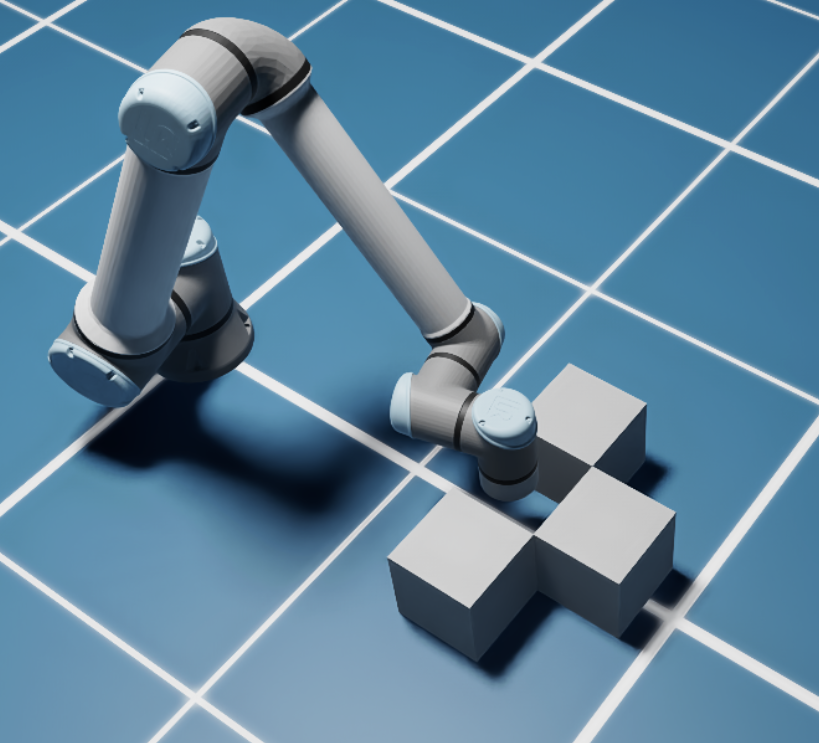}
        \caption{Bucket}
    \end{subfigure}\hfill
    \begin{subfigure}{.49\linewidth}
        \includegraphics[width=\linewidth]{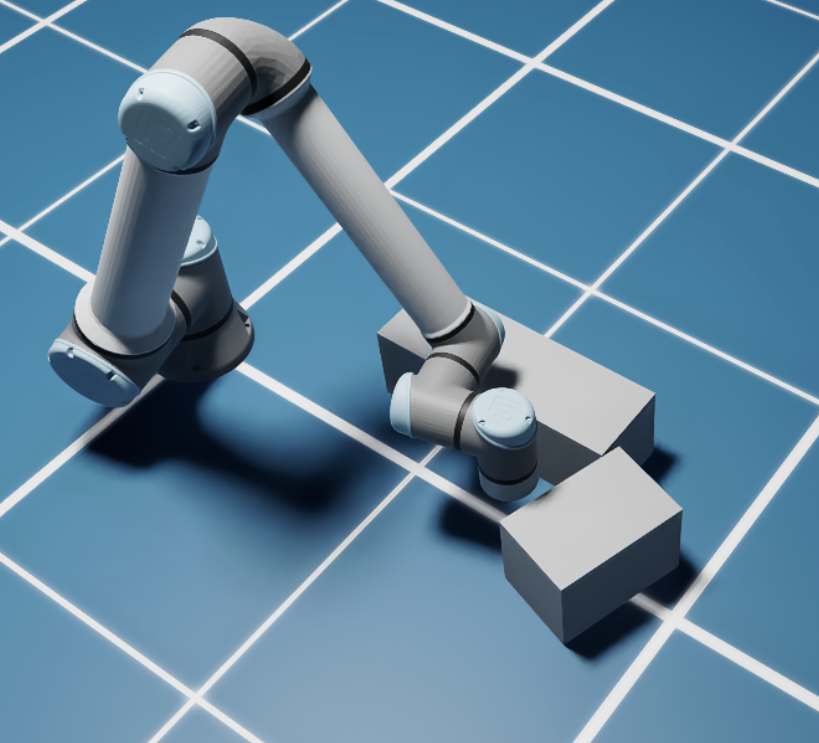}
        \caption{Elbow}
    \end{subfigure}
    \caption{Our approach is evaluated on four experiment setups that test long-term memory, reasoning, and performance.}
    \label{fig:setups}
    \vspace{-4mm}
\end{figure} 

We compare our approach with the baselines on several obstacle configurations (where all obstacles are fixed in place), spanning both simple and complex tasks (see \textref{fig:setups}{Fig.}): \\
\begin{enumerate}
    \item \textbf{Clear}: No obstacles. This setup tests the controllers' ability to directly head towards the goal without unnecessary deviations. 
    \item \textbf{Wall}: A long \SI{45}{\centi\meter} length wall directly in the way. This setup tests the controllers' ability to navigate around standard convex obstacles.
    \item \textbf{Bucket}: Three \SI{15}{\centi\meter} cubes set up to create a concave obstacle directly in the way. This setup tests the ability of the controllers to navigate out of and around simple concave obstacles.
    \item \textbf{Elbow}: A slanted \SI{22.5}{\centi\meter} wall directly in the way redirecting the arm towards a long \SI{45}{\centi\meter} vertical wall. This setup tests the long-term memory of the controllers as they must recall the bottom of the elbow once they cannot pass the vertical section. 
\end{enumerate}

\subsection{Simulation} \label{sec:sim_results}
The simulation experiments were done in NVIDIA's Isaac Sim environment. Each of the four experimental setups (see \textref{fig:setups}{Fig.}) are run 35 times with our approach and the four baselines. The experiment succeeds and terminates if the end-effector of the UR10e moves within \SI{5}{\centi\meter} of the goal position. If it has not succeeded after 1000 iterations of the control loop, the experiment is considered failed. 

In our experiments, every approach succeeded in the \textit{clear} and \textit{wall} tasks every time (see \textref{fig:success_sim}{Fig.}). However, in the \textit{bucket} task, our approach succeeds 85.7\% of the time, occasionally getting stuck on walls. This behaviour may be due to insufficient data or hyperparameter tuning. The baselines with low observation horizons ($H_o=3,6$) fail most \textit{bucket} trials, occasionally succeeding due to their highly stochastic movements. The baselines with $H_o=20,50$ succeed every time. Finally, on the \textit{elbow} task, our approach succeeds 80\% while none of the baselines succeed. This is because the long vertical wall takes longer to explore than the longest observation horizon. Thus, once the baselines have reached the top of the vertical wall, they have forgotten the bottom of the elbow. This causes the baselines to infinitely loop around the inside of the elbow. However, our approach recalls the bottom wall of the elbow and moves around it once it realizes the vertical wall is not traversable. The failures of our approach are due to slow movement or idling near the top of the elbow, causing it to fail to reach the end-point in under 1000 iterations. The idling and slow movement are likely due to the lack of training data in that region; the \textit{elbow} task is outside of the distribution of the training data and causes the end-effector to move closer to the base of the UR10e than the trajectories in training. 

Additionally, our approach almost always reduces the time to completion of the experiments compared to the baselines (see \textref{fig:iters_sim}{Fig.}). In the \textit{clear} setup, our approach moves directly towards the goal while the baselines move in circuitous routes (see \textref{sec:disc}{Section}). All approaches are roughly equal in the \textit{wall} setup. However, in the \textit{bucket} setup, the baselines with low observation horizons (i.e., 3 and 6) have very long convergence times in the few trials where they escape the bucket. Their escape from the bucket is essentially due to random motions, as they do not have enough observation horizon to realize they are in a concave region. With higher observation horizons, the time to completion significantly reduces; however, our approach is still faster than all baselines. In the \textit{elbow} example, none of the baselines succeed, while ours succeeds in roughly the same time required to navigate around the \textit{bucket} setup. 

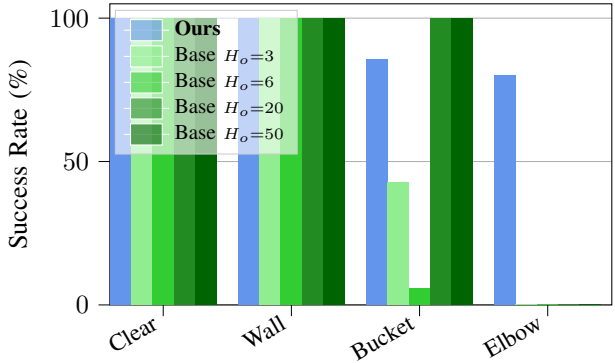
\begin{figure}[t]
    \centering
    \vspace{2mm}
    \input{figs/done_sim}\\
    \vspace{-1mm}
    \caption{The percentage of successful trials for each simulated experiment setup for our approach and the baselines.}
    \label{fig:success_sim}
    \vspace{-5mm}
\end{figure} 

\begin{figure}[t]
    \centering
    \vspace{2mm}
    \input{figs/iters_sim}\\
    \vspace{-3mm}
    \caption{The time required to complete each simulated experiment setup for our approach and the baselines. Only successful trials are plotted. }
    \label{fig:iters_sim}
\end{figure}
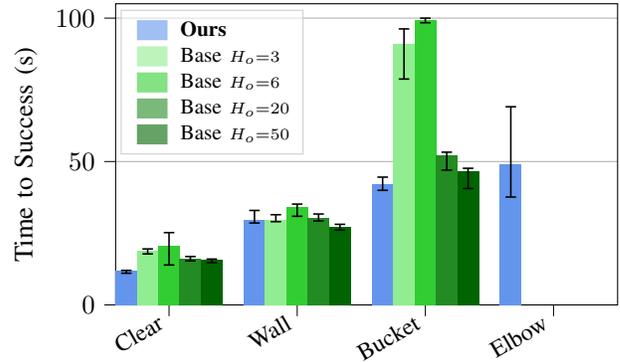

\subsection{Real Experiments} \label{sec:real_results}
Real experiments were conducted with a UR10e with a cylindrical end-effector. The end-effector was 3D printed to increase sensitivity to torque readings and provide additional safety for the real UR10e. The obstacles were rigid cardboard boxes fixed to a flat surface. The real setup had identical positioning of the obstacles to the simulation setup. To prevent damage to the arm, the applied force was limited when the UR10e sensed high torque. The simulation and real experiment platforms returned similar results, demonstrating diffusion's robustness to changes in the environment. 

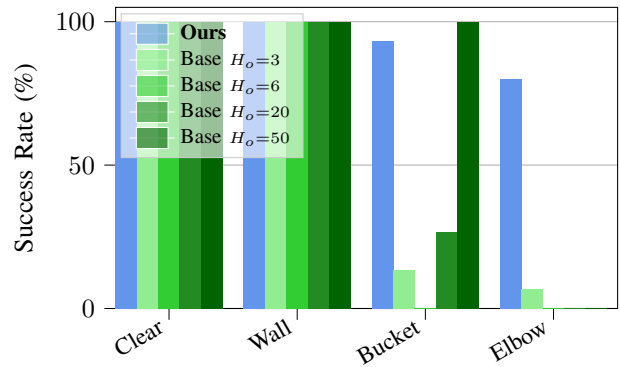
\begin{figure}[t]
    \centering
    \input{figs/done_real}\\
    \caption{The percentage of successful trials for each real experiment setup for our approach and the baselines.}
    \label{fig:success_real}
    \vspace{-4mm}
\end{figure} 

\begin{figure}[t]
    \centering
    \vspace{2mm}
    \input{figs/iters_real}\\
    \caption{The time required to complete each real experiment setup for our approach and the baselines. Only successful trials are plotted.}
    \label{fig:iters_real}
    \vspace{-2mm}
\end{figure}
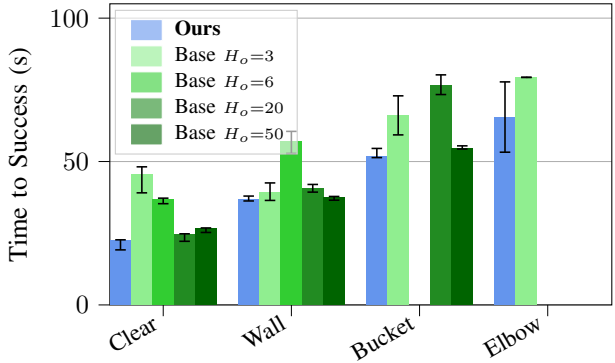 

The same experiments were repeated on the real setup with 15 trials per setup. Similar results were observed and are summarized in \textref{fig:success_real}{Fig.} and \textref{fig:iters_real}{Fig.}. Major differences between the simulation and real results include the significantly worse performance of the baselines with low observation horizons on the \textit{clear} and \textit{wall} setups and the significantly lower success rate of the $H_o=20$ baseline on the \textit{bucket} task. Additionally, it was observed that our approach would occasionally overshoot the goal and then idle during the \textit{elbow} trials. This may be because the end-effector never overshoots the goal in the training data, and thus, the model does not have sufficient data in that region to backtrack to the goal. 

All approaches generally took longer on the real UR10e. This may have been due to the sim-to-real transfer or the lower reactiveness of the real UR10e. However, our approach completed every experiment setup faster than the baselines.

Finally, we measure the mean time required for the diffusion model to complete an inference step (see \textref{fig:comp_sim}{Fig.}). This was determined from the real experiments run using an Intel NUC Extreme, with a 12th generation Intel Core i9-12900 and a GeForce RTX 3080 Ti NVidia graphics card. Our approach has a similar inference time to the $H_o=3,6$ baselines, but as the observation horizon increases, the computation time increases well past our approach. It was not feasible to test with observation horizons larger than 50 due to the inference time taking significantly longer than \SI{0.1}{\second} as required for \SI{10}{\hertz} real-time control. 

\begin{figure}[t]
  \centering
  \input{figs/time_real}\\
  \caption{The time per diffusion inference step for our approach and the baselines.}
  \label{fig:comp_sim}
  \vspace{-5mm}
\end{figure}
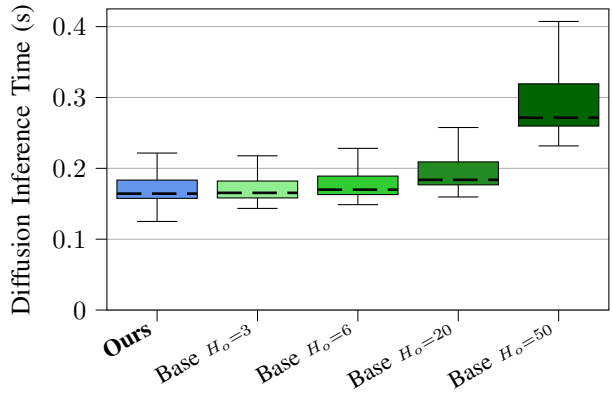

\subsection{Discussion} \label{sec:disc}
The baselines did not move directly toward the goal. Instead, they drift diagonally right until blocked by an obstacle, then perform wall-following away from the goal until the obstacle has been passed. This behaviour is most evident in the \textit{clear} setup where, despite the absence of obstacles, the baselines take significantly longer than our approach to reach the goal position due to their diagonal movement. This behaviour is more pronounced with lower observation horizons.

Despite this non-optimal behaviour not being in the training data, it was likely learned because the baselines do not have sufficient observations to replicate the behaviour of the training data; without some long-term knowledge of previous collisions, the baselines cannot bypass even the simplest obstacles without introducing some bias, such as moving right. However, with keypoints, our approach displays no bias towards moving right, directly moving towards the goal. With the keypoints, our approach can effectively imitate the behaviour of the human operator without relying on heuristics. The failure of these heuristics is again evident in the \textit{elbow} setup as moving right will not succeed, as the vertical component of the elbow is too long to be traversable from the right. Although this obstacle setup is not in the training dataset, our approach effectively generalizes to this new case. In contrast, the baselines become stuck in an infinite loop wall-following away from the goal, realizing it is not traversable from the right and moving back down to the bottom of the elbow.

\section{Limitations and Future Work}
In our approach, a hard-coded keypoint manager determines which information is retained and used. This keypoint manager is separately engineered from the diffusion network and contains specific hyperparameters that must be tuned for the specific experiment. A future direction for this research is directly incorporating memory management into the diffusion network. This new network architecture would reduce hard-coded engineering work and improve performance as the network can optimize memory management for the unique circumstances of the experiment. Additionally, our approach should be compared to machine learning methods that implicitly remember long-term dependencies, such as RetNet~\cite{retnet} and Mamba~\cite{mamba}. However, these approaches are not generative and scale poorly to long sequences (in the case of transformers) or require continuous dynamics rather than sparse contact events (in the case of structured state-space models)~\cite{mamba}. Our explicit memory system can also be tested with other learning-based controllers.

This work could be extended to maintain information beyond contact points. Object properties determined from proprioception, such as compliance or texture, can be leveraged for navigation~\cite{cluttered}, safety~\cite{skin}, and grasping~\cite{grasping}. Waypoints and other key observations can be recorded as keypoints to expand the applicability of keypoints beyond proprioception. 

\section{Conclusion}
This paper proposes adding long-term memory to learning-based controllers by recording and managing informative keypoints. To enable proprioceptive navigation, a record of past contacts is kept and passed to a diffusion model, which determines appropriate actions to take in light of this information. This eliminates the need for explicit mapping and the assumptions tied to mapping. This capability is necessary for robotic systems that rely mostly or wholly on proprioceptive sensors. Robots can use this approach to robustly operate in dangerous and occluded environments, including space and underwater exploration and construction. 

\section*{Acknowledgement}
The research was carried out at the Jet Propulsion Laboratory, California Institute of Technology, under a contract with the National Aeronautics and Space Administration (80NM0018D0004). The high-performance computing resources used in this investigation were provided by funding from the Jet Propulsion Laboratory~(JPL) Information and Technology Solutions Directorate. 

\balance
\bibliographystyle{IEEEtran} 
\bibliography{bibliography}
\balance
\end{document}

%% file: figs/done_sim.tex
\begin{tikzpicture}

\definecolor{cornflowerblue}{RGB}{100,149,237}
\definecolor{darkgray176}{RGB}{176,176,176}
\definecolor{darkgreen}{RGB}{0,100,0}
\definecolor{forestgreen}{RGB}{34,139,34}
\definecolor{lightgreen}{RGB}{144,238,144}
\definecolor{limegreen}{RGB}{50,205,50}
\definecolor{lightgray204}{RGB}{204,204,204}

\pgfdeclareplotmark{ours}
{%
    \path[fill=cornflowerblue] (-3\pgfplotmarksize,-2\pgfplotmarksize) rectangle (3\pgfplotmarksize,2\pgfplotmarksize);
}
\pgfdeclareplotmark{b3}
{%
    \path[fill=lightgreen] (-3\pgfplotmarksize,-2\pgfplotmarksize) rectangle (3\pgfplotmarksize,2\pgfplotmarksize);
}
\pgfdeclareplotmark{b6}
{%
    \path[fill=limegreen] (-3\pgfplotmarksize,-2\pgfplotmarksize) rectangle (3\pgfplotmarksize,2\pgfplotmarksize);
}
\pgfdeclareplotmark{b20}
{%
    \path[fill=forestgreen] (-3\pgfplotmarksize,-2\pgfplotmarksize) rectangle (3\pgfplotmarksize,2\pgfplotmarksize);
}
\pgfdeclareplotmark{b50}
{%
    \path[fill=darkgreen] (-3\pgfplotmarksize,-2\pgfplotmarksize) rectangle (3\pgfplotmarksize,2\pgfplotmarksize);
}

\begin{axis}[
height=2.2in,
legend cell align={left},
legend style={
  fill opacity=0.6,
  draw opacity=1,
  text opacity=1,
  at={(0.01,0.98)},
  anchor=north west,
  draw=lightgray204,
  font=\footnotesize,
  row sep=-1,
},
tick align=outside,
tick pos=left,
width=3.25in,
x grid style={darkgray176},
xmin=-0.208333333333333, xmax=1.75,
xtick style={color=black},
xtick={0,0.5,1,1.5},
xticklabel style={font=\small, rotate=30.0,anchor=east},
xticklabels={Clear,Wall,Bucket, Elbow},
y grid style={darkgray176},
ylabel={Success Rate (\%)},
ymajorgrids,
ymin=0, ymax=105,
ytick pos=left,
ytick style={color=black}
]
\addlegendimage{mark=ours,white}
\addlegendentry{\textbf{Ours}}
\addlegendimage{mark=b3,white}
\addlegendentry{Base $\scriptstyle H_o = 3$}
\addlegendimage{mark=b6,white}
\addlegendentry{Base $\scriptstyle H_o = 6$}
\addlegendimage{mark=b20,white}
\addlegendentry{Base $\scriptstyle H_o = 20$}
\addlegendimage{mark=b50,white}
\addlegendentry{Base $\scriptstyle H_o = 50$}
\draw[draw=none,fill=cornflowerblue] (axis cs:-0.208333333333333,0) rectangle (axis cs:-0.125,100);
\draw[draw=none,fill=cornflowerblue] (axis cs:0.291666666666667,0) rectangle (axis cs:0.375,100);
\draw[draw=none,fill=cornflowerblue] (axis cs:0.791666666666667,0) rectangle (axis cs:0.875,85.7142857142857);
\draw[draw=none,fill=cornflowerblue] (axis cs:1.29166666666667,0) rectangle (axis cs:1.375,80);
\draw[draw=none,fill=lightgreen] (axis cs:-0.125,0) rectangle (axis cs:-0.0416666666666667,100);
\draw[draw=none,fill=lightgreen] (axis cs:0.375,0) rectangle (axis cs:0.458333333333333,100);
\draw[draw=none,fill=lightgreen] (axis cs:0.875,0) rectangle (axis cs:0.958333333333333,42.8571428571429);
\draw[draw=none,fill=lightgreen] (axis cs:1.375,0) rectangle (axis cs:1.45833333333333,0);
\draw[draw=none,fill=limegreen] (axis cs:-0.0416666666666667,0) rectangle (axis cs:0.0416666666666667,100);
\draw[draw=none,fill=limegreen] (axis cs:0.458333333333333,0) rectangle (axis cs:0.541666666666667,100);
\draw[draw=none,fill=limegreen] (axis cs:0.958333333333333,0) rectangle (axis cs:1.04166666666667,5.71428571428571);
\draw[draw=none,fill=limegreen] (axis cs:1.45833333333333,0) rectangle (axis cs:1.54166666666667,0);
\draw[draw=none,fill=forestgreen] (axis cs:0.0416666666666667,0) rectangle (axis cs:0.125,100);
\draw[draw=none,fill=forestgreen] (axis cs:0.541666666666667,0) rectangle (axis cs:0.625,100);
\draw[draw=none,fill=forestgreen] (axis cs:1.04166666666667,0) rectangle (axis cs:1.125,100);
\draw[draw=none,fill=forestgreen] (axis cs:1.54166666666667,0) rectangle (axis cs:1.625,0);
\draw[draw=none,fill=darkgreen] (axis cs:0.125,0) rectangle (axis cs:0.208333333333333,100);
\draw[draw=none,fill=darkgreen] (axis cs:0.625,0) rectangle (axis cs:0.708333333333333,100);
\draw[draw=none,fill=darkgreen] (axis cs:1.125,0) rectangle (axis cs:1.20833333333333,100);
\draw[draw=none,fill=darkgreen] (axis cs:1.625,0) rectangle (axis cs:1.70833333333333,0);
\end{axis}

\end{tikzpicture}

%% file: figs/iters_sim.tex
\begin{tikzpicture}

\definecolor{cornflowerblue}{RGB}{100,149,237}
\definecolor{darkgray176}{RGB}{176,176,176}
\definecolor{darkgreen}{RGB}{0,100,0}
\definecolor{forestgreen}{RGB}{34,139,34}
\definecolor{lightgreen}{RGB}{144,238,144}
\definecolor{limegreen}{RGB}{50,205,50}
\definecolor{lightgray204}{RGB}{204,204,204}

\pgfdeclareplotmark{ours}
{%
    \path[fill=cornflowerblue] (-3\pgfplotmarksize,-2\pgfplotmarksize) rectangle (3\pgfplotmarksize,2\pgfplotmarksize);
}
\pgfdeclareplotmark{b3}
{%
    \path[fill=lightgreen] (-3\pgfplotmarksize,-2\pgfplotmarksize) rectangle (3\pgfplotmarksize,2\pgfplotmarksize);
}
\pgfdeclareplotmark{b6}
{%
    \path[fill=limegreen] (-3\pgfplotmarksize,-2\pgfplotmarksize) rectangle (3\pgfplotmarksize,2\pgfplotmarksize);
}
\pgfdeclareplotmark{b20}
{%
    \path[fill=forestgreen] (-3\pgfplotmarksize,-2\pgfplotmarksize) rectangle (3\pgfplotmarksize,2\pgfplotmarksize);
}
\pgfdeclareplotmark{b50}
{%
    \path[fill=darkgreen] (-3\pgfplotmarksize,-2\pgfplotmarksize) rectangle (3\pgfplotmarksize,2\pgfplotmarksize);
}

\begin{axis}[
height=2.2in,
legend cell align={left},
legend style={
  fill opacity=0.6,
  draw opacity=1,
  text opacity=1,
  at={(0.01,0.98)},
  anchor=north west,
  draw=lightgray204,
  font=\footnotesize,
  row sep=-1,
},
tick align=outside,
tick pos=left,
width=3.25in,
x grid style={darkgray176},
xmin=-0.208333333333333, xmax=1.75,
xtick style={color=black},
xtick={0,0.5,1,1.5},
xticklabel style={font=\small, rotate=30.0,anchor=east},
xticklabels={Clear,Wall,Bucket,Elbow},
y grid style={darkgray176},
ylabel={Time to Success (s)},
ymajorgrids,
ymin=0, ymax=105,
ytick pos=left,
ytick style={color=black}
]
\addlegendimage{mark=ours,white}
\addlegendentry{\textbf{Ours}}
\addlegendimage{mark=b3,white}
\addlegendentry{Base $\scriptstyle H_o = 3$}
\addlegendimage{mark=b6,white}
\addlegendentry{Base $\scriptstyle H_o = 6$}
\addlegendimage{mark=b20,white}
\addlegendentry{Base $\scriptstyle H_o = 20$}
\addlegendimage{mark=b50,white}
\addlegendentry{Base $\scriptstyle H_o = 50$}
\draw[draw=none,fill=cornflowerblue] (axis cs:-0.208333333333333,0) rectangle (axis cs:-0.125,11.9940640926361);
\draw[draw=none,fill=cornflowerblue] (axis cs:0.291666666666667,0) rectangle (axis cs:0.375,29.804979801178);
\draw[draw=none,fill=cornflowerblue] (axis cs:0.791666666666667,0) rectangle (axis cs:0.875,42.0546146631241);
\draw[draw=none,fill=cornflowerblue] (axis cs:1.29166666666667,0) rectangle (axis cs:1.375,49.1351293325424);
\draw[draw=none,fill=lightgreen] (axis cs:-0.125,0) rectangle (axis cs:-0.0416666666666667,18.9124426841736);
\draw[draw=none,fill=lightgreen] (axis cs:0.375,0) rectangle (axis cs:0.458333333333333,29.7879168987274);
\draw[draw=none,fill=lightgreen] (axis cs:0.875,0) rectangle (axis cs:0.958333333333333,91.0740926265717);
\draw[draw=none,fill=limegreen] (axis cs:-0.0416666666666667,0) rectangle (axis cs:0.0416666666666667,20.6261191368103);
\draw[draw=none,fill=limegreen] (axis cs:0.458333333333333,0) rectangle (axis cs:0.541666666666667,34.254332780838);
\draw[draw=none,fill=limegreen] (axis cs:0.958333333333333,0) rectangle (axis cs:1.04166666666667,99.1988890171051);
\draw[draw=none,fill=forestgreen] (axis cs:0.0416666666666667,0) rectangle (axis cs:0.125,16.1329298019409);
\draw[draw=none,fill=forestgreen] (axis cs:0.541666666666667,0) rectangle (axis cs:0.625,30.4431521892548);
\draw[draw=none,fill=forestgreen] (axis cs:1.04166666666667,0) rectangle (axis cs:1.125,52.1244382858276);
\draw[draw=none,fill=darkgreen] (axis cs:0.125,0) rectangle (axis cs:0.208333333333333,15.9349961280823);
\draw[draw=none,fill=darkgreen] (axis cs:0.625,0) rectangle (axis cs:0.708333333333333,27.3040771484375);
\draw[draw=none,fill=darkgreen] (axis cs:1.125,0) rectangle (axis cs:1.20833333333333,46.7122051715851);
\path [draw=black, semithick]
(axis cs:-0.166666666666667,11.0608596801758)
--(axis cs:-0.166666666666667,12.040629863739);

\addplot [semithick, black, mark=-, mark size=2, mark options={solid}, only marks]
table {%
-0.166666666666667 11.0608596801758
};
\addplot [semithick, black, mark=-, mark size=2, mark options={solid}, only marks]
table {%
-0.166666666666667 12.040629863739
};
\path [draw=black, semithick]
(axis cs:0.333333333333333,28.5686888694763)
--(axis cs:0.333333333333333,32.9435222148895);

\addplot [semithick, black, mark=-, mark size=2, mark options={solid}, only marks]
table {%
0.333333333333333 28.5686888694763
};
\addplot [semithick, black, mark=-, mark size=2, mark options={solid}, only marks]
table {%
0.333333333333333 32.9435222148895
};
\path [draw=black, semithick]
(axis cs:0.833333333333333,39.9590361118317)
--(axis cs:0.833333333333333,44.5637881755829);

\addplot [semithick, black, mark=-, mark size=2, mark options={solid}, only marks]
table {%
0.833333333333333 39.9590361118317
};
\addplot [semithick, black, mark=-, mark size=2, mark options={solid}, only marks]
table {%
0.833333333333333 44.5637881755829
};
\path [draw=black, semithick]
(axis cs:1.33333333333333,37.6214913129807)
--(axis cs:1.33333333333333,69.0961608886719);

\addplot [semithick, black, mark=-, mark size=2, mark options={solid}, only marks]
table {%
1.33333333333333 37.6214913129807
};
\addplot [semithick, black, mark=-, mark size=2, mark options={solid}, only marks]
table {%
1.33333333333333 69.0961608886719
};
\path [draw=black, semithick]
(axis cs:-0.0833333333333333,17.793560385704)
--(axis cs:-0.0833333333333333,19.5192939043045);

\addplot [semithick, black, mark=-, mark size=2, mark options={solid}, only marks]
table {%
-0.0833333333333333 17.793560385704
};
\addplot [semithick, black, mark=-, mark size=2, mark options={solid}, only marks]
table {%
-0.0833333333333333 19.5192939043045
};
\path [draw=black, semithick]
(axis cs:0.416666666666667,29.0257551670074)
--(axis cs:0.416666666666667,31.4594215154648);

\addplot [semithick, black, mark=-, mark size=2, mark options={solid}, only marks]
table {%
0.416666666666667 29.0257551670074
};
\addplot [semithick, black, mark=-, mark size=2, mark options={solid}, only marks]
table {%
0.416666666666667 31.4594215154648
};
\path [draw=black, semithick]
(axis cs:0.916666666666667,78.7563672065735)
--(axis cs:0.916666666666667,96.2466484308243);

\addplot [semithick, black, mark=-, mark size=2, mark options={solid}, only marks]
table {%
0.916666666666667 78.7563672065735
};
\addplot [semithick, black, mark=-, mark size=2, mark options={solid}, only marks]
table {%
0.916666666666667 96.2466484308243
};
\path [draw=black, semithick]
(axis cs:0,13.9238270521164)
--(axis cs:0,25.2154139280319);

\addplot [semithick, black, mark=-, mark size=2, mark options={solid}, only marks]
table {%
0 13.9238270521164
};
\addplot [semithick, black, mark=-, mark size=2, mark options={solid}, only marks]
table {%
0 25.2154139280319
};
\path [draw=black, semithick]
(axis cs:0.5,30.9036864042282)
--(axis cs:0.5,35.1608955860138);

\addplot [semithick, black, mark=-, mark size=2, mark options={solid}, only marks]
table {%
0.5 30.9036864042282
};
\addplot [semithick, black, mark=-, mark size=2, mark options={solid}, only marks]
table {%
0.5 35.1608955860138
};
\path [draw=black, semithick]
(axis cs:1,98.4030128717422)
--(axis cs:1,99.994765162468);

\addplot [semithick, black, mark=-, mark size=2, mark options={solid}, only marks]
table {%
1 98.4030128717422
};
\addplot [semithick, black, mark=-, mark size=2, mark options={solid}, only marks]
table {%
1 99.994765162468
};
\path [draw=black, semithick]
(axis cs:0.0833333333333333,15.3447337150574)
--(axis cs:0.0833333333333333,16.8662339448929);

\addplot [semithick, black, mark=-, mark size=2, mark options={solid}, only marks]
table {%
0.0833333333333333 15.3447337150574
};
\addplot [semithick, black, mark=-, mark size=2, mark options={solid}, only marks]
table {%
0.0833333333333333 16.8662339448929
};
\path [draw=black, semithick]
(axis cs:0.583333333333333,29.2866864204407)
--(axis cs:0.583333333333333,31.7000014781952);

\addplot [semithick, black, mark=-, mark size=2, mark options={solid}, only marks]
table {%
0.583333333333333 29.2866864204407
};
\addplot [semithick, black, mark=-, mark size=2, mark options={solid}, only marks]
table {%
0.583333333333333 31.7000014781952
};
\path [draw=black, semithick]
(axis cs:1.08333333333333,47.0147042274475)
--(axis cs:1.08333333333333,53.2614623308182);

\addplot [semithick, black, mark=-, mark size=2, mark options={solid}, only marks]
table {%
1.08333333333333 47.0147042274475
};
\addplot [semithick, black, mark=-, mark size=2, mark options={solid}, only marks]
table {%
1.08333333333333 53.2614623308182
};
\path [draw=black, semithick]
(axis cs:0.166666666666667,14.7718917131424)
--(axis cs:0.166666666666667,15.9875295162201);

\addplot [semithick, black, mark=-, mark size=2, mark options={solid}, only marks]
table {%
0.166666666666667 14.7718917131424
};
\addplot [semithick, black, mark=-, mark size=2, mark options={solid}, only marks]
table {%
0.166666666666667 15.9875295162201
};
\path [draw=black, semithick]
(axis cs:0.666666666666667,26.0834736824036)
--(axis cs:0.666666666666667,28.1080045700073);

\addplot [semithick, black, mark=-, mark size=2, mark options={solid}, only marks]
table {%
0.666666666666667 26.0834736824036
};
\addplot [semithick, black, mark=-, mark size=2, mark options={solid}, only marks]
table {%
0.666666666666667 28.1080045700073
};
\path [draw=black, semithick]
(axis cs:1.16666666666667,40.5830833911896)
--(axis cs:1.16666666666667,47.6507844924927);

\addplot [semithick, black, mark=-, mark size=2, mark options={solid}, only marks]
table {%
1.16666666666667 40.5830833911896
};
\addplot [semithick, black, mark=-, mark size=2, mark options={solid}, only marks]
table {%
1.16666666666667 47.6507844924927
};
\addplot [semithick, black]
table {%
-0.166666666666667 11.9940640926361
};
\addplot [semithick, black]
table {%
0.333333333333333 29.804979801178
};
\addplot [semithick, black]
table {%
0.833333333333333 42.0546146631241
};
\addplot [semithick, black]
table {%
1.33333333333333 49.1351293325424
};
\addplot [semithick, black]
table {%
-0.0833333333333333 18.9124426841736
};
\addplot [semithick, black]
table {%
0.416666666666667 29.7879168987274
};
\addplot [semithick, black]
table {%
0.916666666666667 91.0740926265717
};
\addplot [semithick, black]
table {%
0 20.6261191368103
};
\addplot [semithick, black]
table {%
0.5 34.254332780838
};
\addplot [semithick, black]
table {%
1 99.1988890171051
};
\addplot [semithick, black]
table {%
0.0833333333333333 16.1329298019409
};
\addplot [semithick, black]
table {%
0.583333333333333 30.4431521892548
};
\addplot [semithick, black]
table {%
1.08333333333333 52.1244382858276
};
\addplot [semithick, black]
table {%
0.166666666666667 15.9349961280823
};
\addplot [semithick, black]
table {%
0.666666666666667 27.3040771484375
};
\addplot [semithick, black]
table {%
1.16666666666667 46.7122051715851
};
\end{axis}

\end{tikzpicture}

%% file: figs/done_real.tex
\begin{tikzpicture}

\definecolor{cornflowerblue}{RGB}{100,149,237}
\definecolor{darkgray176}{RGB}{176,176,176}
\definecolor{darkgreen}{RGB}{0,100,0}
\definecolor{forestgreen}{RGB}{34,139,34}
\definecolor{lightgreen}{RGB}{144,238,144}
\definecolor{limegreen}{RGB}{50,205,50}
\definecolor{lightgray204}{RGB}{204,204,204}

\pgfdeclareplotmark{ours}
{%
    \path[fill=cornflowerblue] (-3\pgfplotmarksize,-2\pgfplotmarksize) rectangle (3\pgfplotmarksize,2\pgfplotmarksize);
}
\pgfdeclareplotmark{b3}
{%
    \path[fill=lightgreen] (-3\pgfplotmarksize,-2\pgfplotmarksize) rectangle (3\pgfplotmarksize,2\pgfplotmarksize);
}
\pgfdeclareplotmark{b6}
{%
    \path[fill=limegreen] (-3\pgfplotmarksize,-2\pgfplotmarksize) rectangle (3\pgfplotmarksize,2\pgfplotmarksize);
}
\pgfdeclareplotmark{b20}
{%
    \path[fill=forestgreen] (-3\pgfplotmarksize,-2\pgfplotmarksize) rectangle (3\pgfplotmarksize,2\pgfplotmarksize);
}
\pgfdeclareplotmark{b50}
{%
    \path[fill=darkgreen] (-3\pgfplotmarksize,-2\pgfplotmarksize) rectangle (3\pgfplotmarksize,2\pgfplotmarksize);
}

\begin{axis}[
height=2.2in,
legend cell align={left},
legend style={
  fill opacity=0.6,
  draw opacity=1,
  text opacity=1,
  at={(0.01,0.98)},
  anchor=north west,
  draw=lightgray204,
  font=\footnotesize,
  row sep=-1,
},
tick align=outside,
tick pos=left,
width=3.25in,
x grid style={darkgray176},
xmin=-0.208333333333333, xmax=1.75,
xtick style={color=black},
xtick={0,0.5,1,1.5},
xticklabel style={font=\small, rotate=30.0,anchor=east},
xticklabels={Clear,Wall,Bucket,Elbow},
y grid style={darkgray176},
ylabel={Success Rate (\%)},
ymajorgrids,
ymin=0, ymax=105,
ytick pos=left,
ytick style={color=black}
]
\addlegendimage{mark=ours,white}
\addlegendentry{\textbf{Ours}}
\addlegendimage{mark=b3,white}
\addlegendentry{Base $\scriptstyle H_o = 3$}
\addlegendimage{mark=b6,white}
\addlegendentry{Base $\scriptstyle H_o = 6$}
\addlegendimage{mark=b20,white}
\addlegendentry{Base $\scriptstyle H_o = 20$}
\addlegendimage{mark=b50,white}
\addlegendentry{Base $\scriptstyle H_o = 50$}
\draw[draw=none,fill=cornflowerblue] (axis cs:-0.208333333333333,0) rectangle (axis cs:-0.125,100);
\draw[draw=none,fill=cornflowerblue] (axis cs:0.291666666666667,0) rectangle (axis cs:0.375,100);
\draw[draw=none,fill=cornflowerblue] (axis cs:0.791666666666667,0) rectangle (axis cs:0.875,93.3333333333333);
\draw[draw=none,fill=cornflowerblue] (axis cs:1.29166666666667,0) rectangle (axis cs:1.375,80);
\draw[draw=none,fill=lightgreen] (axis cs:-0.125,0) rectangle (axis cs:-0.0416666666666667,100);
\draw[draw=none,fill=lightgreen] (axis cs:0.375,0) rectangle (axis cs:0.458333333333333,100);
\draw[draw=none,fill=lightgreen] (axis cs:0.875,0) rectangle (axis cs:0.958333333333333,13.3333333333333);
\draw[draw=none,fill=lightgreen] (axis cs:1.375,0) rectangle (axis cs:1.45833333333333,6.66666666666667);
\draw[draw=none,fill=limegreen] (axis cs:-0.0416666666666667,0) rectangle (axis cs:0.0416666666666667,100);
\draw[draw=none,fill=limegreen] (axis cs:0.458333333333333,0) rectangle (axis cs:0.541666666666667,100);
\draw[draw=none,fill=limegreen] (axis cs:0.958333333333333,0) rectangle (axis cs:1.04166666666667,0);
\draw[draw=none,fill=limegreen] (axis cs:1.45833333333333,0) rectangle (axis cs:1.54166666666667,0);
\draw[draw=none,fill=forestgreen] (axis cs:0.0416666666666667,0) rectangle (axis cs:0.125,100);
\draw[draw=none,fill=forestgreen] (axis cs:0.541666666666667,0) rectangle (axis cs:0.625,100);
\draw[draw=none,fill=forestgreen] (axis cs:1.04166666666667,0) rectangle (axis cs:1.125,26.6666666666667);
\draw[draw=none,fill=forestgreen] (axis cs:1.54166666666667,0) rectangle (axis cs:1.625,0);
\draw[draw=none,fill=darkgreen] (axis cs:0.125,0) rectangle (axis cs:0.208333333333333,100);
\draw[draw=none,fill=darkgreen] (axis cs:0.625,0) rectangle (axis cs:0.708333333333333,100);
\draw[draw=none,fill=darkgreen] (axis cs:1.125,0) rectangle (axis cs:1.20833333333333,100);
\draw[draw=none,fill=darkgreen] (axis cs:1.625,0) rectangle (axis cs:1.70833333333333,0);
\end{axis}

\end{tikzpicture}

%% file: figs/iters_real.tex
\begin{tikzpicture}

\definecolor{cornflowerblue}{RGB}{100,149,237}
\definecolor{darkgray176}{RGB}{176,176,176}
\definecolor{darkgreen}{RGB}{0,100,0}
\definecolor{forestgreen}{RGB}{34,139,34}
\definecolor{lightgreen}{RGB}{144,238,144}
\definecolor{limegreen}{RGB}{50,205,50}
\definecolor{lightgray204}{RGB}{204,204,204}

\pgfdeclareplotmark{ours}
{%
    \path[fill=cornflowerblue] (-3\pgfplotmarksize,-2\pgfplotmarksize) rectangle (3\pgfplotmarksize,2\pgfplotmarksize);
}
\pgfdeclareplotmark{b3}
{%
    \path[fill=lightgreen] (-3\pgfplotmarksize,-2\pgfplotmarksize) rectangle (3\pgfplotmarksize,2\pgfplotmarksize);
}
\pgfdeclareplotmark{b6}
{%
    \path[fill=limegreen] (-3\pgfplotmarksize,-2\pgfplotmarksize) rectangle (3\pgfplotmarksize,2\pgfplotmarksize);
}
\pgfdeclareplotmark{b20}
{%
    \path[fill=forestgreen] (-3\pgfplotmarksize,-2\pgfplotmarksize) rectangle (3\pgfplotmarksize,2\pgfplotmarksize);
}
\pgfdeclareplotmark{b50}
{%
    \path[fill=darkgreen] (-3\pgfplotmarksize,-2\pgfplotmarksize) rectangle (3\pgfplotmarksize,2\pgfplotmarksize);
}

\begin{axis}[
height=2.2in,
legend cell align={left},
legend style={
  fill opacity=0.6,
  draw opacity=1,
  text opacity=1,
  at={(0.01,0.98)},
  anchor=north west,
  draw=lightgray204,
  font=\footnotesize,
  row sep=-1,
},
tick align=outside,
tick pos=left,
width=3.25in,
x grid style={darkgray176},
xmin=-0.208333333333333, xmax=1.75,
xtick style={color=black},
xtick={0,0.5,1,1.5},
xticklabel style={font=\small, rotate=30.0,anchor=east},
xticklabels={Clear,Wall,Bucket,Elbow},
y grid style={darkgray176},
ylabel={Time to Success (s)},
ymajorgrids,
ymin=0, ymax=105,
ytick pos=left,
ytick style={color=black}
]
\addlegendimage{mark=ours,white}
\addlegendentry{\textbf{Ours}}
\addlegendimage{mark=b3,white}
\addlegendentry{Base $\scriptstyle H_o = 3$}
\addlegendimage{mark=b6,white}
\addlegendentry{Base $\scriptstyle H_o = 6$}
\addlegendimage{mark=b20,white}
\addlegendentry{Base $\scriptstyle H_o = 20$}
\addlegendimage{mark=b50,white}
\addlegendentry{Base $\scriptstyle H_o = 50$}
\draw[draw=none,fill=cornflowerblue] (axis cs:-0.208333333333333,0) rectangle (axis cs:-0.125,22.4339377880096);
\draw[draw=none,fill=cornflowerblue] (axis cs:0.291666666666667,0) rectangle (axis cs:0.375,36.8071773052216);
\draw[draw=none,fill=cornflowerblue] (axis cs:0.791666666666667,0) rectangle (axis cs:0.875,51.8157035112381);
\draw[draw=none,fill=cornflowerblue] (axis cs:1.29166666666667,0) rectangle (axis cs:1.375,65.5138592720032);
\draw[draw=none,fill=lightgreen] (axis cs:-0.125,0) rectangle (axis cs:-0.0416666666666667,45.6005218029022);
\draw[draw=none,fill=lightgreen] (axis cs:0.375,0) rectangle (axis cs:0.458333333333333,39.3271255493164);
\draw[draw=none,fill=lightgreen] (axis cs:0.875,0) rectangle (axis cs:0.958333333333333,66.0993223190308);
\draw[draw=none,fill=lightgreen] (axis cs:1.375,0) rectangle (axis cs:1.45833333333333,79.3497884273529);
\draw[draw=none,fill=limegreen] (axis cs:-0.0416666666666667,0) rectangle (axis cs:0.0416666666666667,36.8888685703278);
\draw[draw=none,fill=limegreen] (axis cs:0.458333333333333,0) rectangle (axis cs:0.541666666666667,56.8993403911591);
\draw[draw=none,fill=forestgreen] (axis cs:0.0416666666666667,0) rectangle (axis cs:0.125,24.6711785793304);
\draw[draw=none,fill=forestgreen] (axis cs:0.541666666666667,0) rectangle (axis cs:0.625,40.6567432880402);
\draw[draw=none,fill=forestgreen] (axis cs:1.04166666666667,0) rectangle (axis cs:1.125,76.5974996089935);
\draw[draw=none,fill=darkgreen] (axis cs:0.125,0) rectangle (axis cs:0.208333333333333,26.5161769390106);
\draw[draw=none,fill=darkgreen] (axis cs:0.625,0) rectangle (axis cs:0.708333333333333,37.6023304462433);
\draw[draw=none,fill=darkgreen] (axis cs:1.125,0) rectangle (axis cs:1.20833333333333,54.6238360404968);
\path [draw=black, semithick]
(axis cs:-0.166666666666667,19.1952612400055)
--(axis cs:-0.166666666666667,22.7020517587662);

\addplot [semithick, black, mark=-, mark size=2, mark options={solid}, only marks]
table {%
-0.166666666666667 19.1952612400055
};
\addplot [semithick, black, mark=-, mark size=2, mark options={solid}, only marks]
table {%
-0.166666666666667 22.7020517587662
};
\path [draw=black, semithick]
(axis cs:0.333333333333333,36.2494086027145)
--(axis cs:0.333333333333333,37.9577438831329);

\addplot [semithick, black, mark=-, mark size=2, mark options={solid}, only marks]
table {%
0.333333333333333 36.2494086027145
};
\addplot [semithick, black, mark=-, mark size=2, mark options={solid}, only marks]
table {%
0.333333333333333 37.9577438831329
};
\path [draw=black, semithick]
(axis cs:0.833333333333333,51.3641529679298)
--(axis cs:0.833333333333333,54.5673049688339);

\addplot [semithick, black, mark=-, mark size=2, mark options={solid}, only marks]
table {%
0.833333333333333 51.3641529679298
};
\addplot [semithick, black, mark=-, mark size=2, mark options={solid}, only marks]
table {%
0.833333333333333 54.5673049688339
};
\path [draw=black, semithick]
(axis cs:1.33333333333333,53.2554513812065)
--(axis cs:1.33333333333333,77.7688781619072);

\addplot [semithick, black, mark=-, mark size=2, mark options={solid}, only marks]
table {%
1.33333333333333 53.2554513812065
};
\addplot [semithick, black, mark=-, mark size=2, mark options={solid}, only marks]
table {%
1.33333333333333 77.7688781619072
};
\path [draw=black, semithick]
(axis cs:-0.0833333333333333,39.084327340126)
--(axis cs:-0.0833333333333333,48.1523824930191);

\addplot [semithick, black, mark=-, mark size=2, mark options={solid}, only marks]
table {%
-0.0833333333333333 39.084327340126
};
\addplot [semithick, black, mark=-, mark size=2, mark options={solid}, only marks]
table {%
-0.0833333333333333 48.1523824930191
};
\path [draw=black, semithick]
(axis cs:0.416666666666667,36.4382145404816)
--(axis cs:0.416666666666667,42.5409553050995);

\addplot [semithick, black, mark=-, mark size=2, mark options={solid}, only marks]
table {%
0.416666666666667 36.4382145404816
};
\addplot [semithick, black, mark=-, mark size=2, mark options={solid}, only marks]
table {%
0.416666666666667 42.5409553050995
};
\path [draw=black, semithick]
(axis cs:0.916666666666667,59.2902816534042)
--(axis cs:0.916666666666667,72.9083629846573);

\addplot [semithick, black, mark=-, mark size=2, mark options={solid}, only marks]
table {%
0.916666666666667 59.2902816534042
};
\addplot [semithick, black, mark=-, mark size=2, mark options={solid}, only marks]
table {%
0.916666666666667 72.9083629846573
};
\path [draw=black, semithick]
(axis cs:1.41666666666667,79.3497884273529)
--(axis cs:1.41666666666667,79.3497884273529);

\addplot [semithick, black, mark=-, mark size=2, mark options={solid}, only marks]
table {%
1.41666666666667 79.3497884273529
};
\addplot [semithick, black, mark=-, mark size=2, mark options={solid}, only marks]
table {%
1.41666666666667 79.3497884273529
};
\path [draw=black, semithick]
(axis cs:0,35.3008260726929)
--(axis cs:0,37.2271114587784);

\addplot [semithick, black, mark=-, mark size=2, mark options={solid}, only marks]
table {%
0 35.3008260726929
};
\addplot [semithick, black, mark=-, mark size=2, mark options={solid}, only marks]
table {%
0 37.2271114587784
};
\path [draw=black, semithick]
(axis cs:0.5,52.8302999734879)
--(axis cs:0.5,60.5116472244263);

\addplot [semithick, black, mark=-, mark size=2, mark options={solid}, only marks]
table {%
0.5 52.8302999734879
};
\addplot [semithick, black, mark=-, mark size=2, mark options={solid}, only marks]
table {%
0.5 60.5116472244263
};
\path [draw=black, semithick]
(axis cs:0.0833333333333333,22.1838670969009)
--(axis cs:0.0833333333333333,24.7985700368881);

\addplot [semithick, black, mark=-, mark size=2, mark options={solid}, only marks]
table {%
0.0833333333333333 22.1838670969009
};
\addplot [semithick, black, mark=-, mark size=2, mark options={solid}, only marks]
table {%
0.0833333333333333 24.7985700368881
};
\path [draw=black, semithick]
(axis cs:0.583333333333333,39.3330564498901)
--(axis cs:0.583333333333333,42.011225938797);

\addplot [semithick, black, mark=-, mark size=2, mark options={solid}, only marks]
table {%
0.583333333333333 39.3330564498901
};
\addplot [semithick, black, mark=-, mark size=2, mark options={solid}, only marks]
table {%
0.583333333333333 42.011225938797
};
\path [draw=black, semithick]
(axis cs:1.08333333333333,73.3550786972046)
--(axis cs:1.08333333333333,80.2223613262177);

\addplot [semithick, black, mark=-, mark size=2, mark options={solid}, only marks]
table {%
1.08333333333333 73.3550786972046
};
\addplot [semithick, black, mark=-, mark size=2, mark options={solid}, only marks]
table {%
1.08333333333333 80.2223613262177
};
\path [draw=black, semithick]
(axis cs:0.166666666666667,25.2682952880859)
--(axis cs:0.166666666666667,26.8445771932602);

\addplot [semithick, black, mark=-, mark size=2, mark options={solid}, only marks]
table {%
0.166666666666667 25.2682952880859
};
\addplot [semithick, black, mark=-, mark size=2, mark options={solid}, only marks]
table {%
0.166666666666667 26.8445771932602
};
\path [draw=black, semithick]
(axis cs:0.666666666666667,36.5179980993271)
--(axis cs:0.666666666666667,37.8243832588196);

\addplot [semithick, black, mark=-, mark size=2, mark options={solid}, only marks]
table {%
0.666666666666667 36.5179980993271
};
\addplot [semithick, black, mark=-, mark size=2, mark options={solid}, only marks]
table {%
0.666666666666667 37.8243832588196
};
\path [draw=black, semithick]
(axis cs:1.16666666666667,54.3496395349503)
--(axis cs:1.16666666666667,55.4699280261993);

\addplot [semithick, black, mark=-, mark size=2, mark options={solid}, only marks]
table {%
1.16666666666667 54.3496395349503
};
\addplot [semithick, black, mark=-, mark size=2, mark options={solid}, only marks]
table {%
1.16666666666667 55.4699280261993
};
\addplot [semithick, black]
table {%
-0.166666666666667 22.4339377880096
};
\addplot [semithick, black]
table {%
0.333333333333333 36.8071773052216
};
\addplot [semithick, black]
table {%
0.833333333333333 51.8157035112381
};
\addplot [semithick, black]
table {%
1.33333333333333 65.5138592720032
};
\addplot [semithick, black]
table {%
-0.0833333333333333 45.6005218029022
};
\addplot [semithick, black]
table {%
0.416666666666667 39.3271255493164
};
\addplot [semithick, black]
table {%
0.916666666666667 66.0993223190308
};
\addplot [semithick, black]
table {%
1.41666666666667 79.3497884273529
};
\addplot [semithick, black]
table {%
0 36.8888685703278
};
\addplot [semithick, black]
table {%
0.5 56.8993403911591
};
\addplot [semithick, black]
table {%
0.0833333333333333 24.6711785793304
};
\addplot [semithick, black]
table {%
0.583333333333333 40.6567432880402
};
\addplot [semithick, black]
table {%
1.08333333333333 76.5974996089935
};
\addplot [semithick, black]
table {%
0.166666666666667 26.5161769390106
};
\addplot [semithick, black]
table {%
0.666666666666667 37.6023304462433
};
\addplot [semithick, black]
table {%
1.16666666666667 54.6238360404968
};

\end{axis}

\end{tikzpicture}

%% file: figs/time_real.tex
\begin{tikzpicture}

\definecolor{cornflowerblue}{RGB}{100,149,237}
\definecolor{darkgray176}{RGB}{176,176,176}
\definecolor{darkgreen}{RGB}{0,100,0}
\definecolor{forestgreen}{RGB}{34,139,34}
\definecolor{lightgreen}{RGB}{144,238,144}
\definecolor{limegreen}{RGB}{50,205,50}

\begin{axis}[
height=2.2in,
tick align=outside,
tick pos=left,
width=3.25in,
x grid style={darkgray176},
xmin=-0.5, xmax=4.5,
xtick style={color=black},
xtick={0,1,2,3,4},
xticklabel style={font=\small, rotate=30.0,anchor=east},
xticklabels={\textbf{Ours}, Base $\scriptstyle H_o = 3$, Base $\scriptstyle H_o = 6$, Base $\scriptstyle H_o = 20$, Base $\scriptstyle H_o = 50$},
y grid style={darkgray176},
ylabel={Diffusion Inference Time (s)},
ymajorgrids,
ymin=0.0, ymax=0.425,
ytick pos=left,
ytick style={color=black}
]
\path [draw=black, fill=cornflowerblue]
(axis cs:-0.4,0.157605707645416)
--(axis cs:0.4,0.157605707645416)
--(axis cs:0.4,0.183418810367584)
--(axis cs:-0.4,0.183418810367584)
--(axis cs:-0.4,0.157605707645416)
--cycle;
\addplot [black]
table {%
0 0.157605707645416
0 0.125023603439331
};
\addplot [black]
table {%
0 0.183418810367584
0 0.221582412719727
};
\addplot [black]
table {%
-0.2 0.125023603439331
0.2 0.125023603439331
};
\addplot [black]
table {%
-0.2 0.221582412719727
0.2 0.221582412719727
};
\path [draw=black, fill=lightgreen]
(axis cs:0.6,0.158243477344513)
--(axis cs:1.4,0.158243477344513)
--(axis cs:1.4,0.18207985162735)
--(axis cs:0.6,0.18207985162735)
--(axis cs:0.6,0.158243477344513)
--cycle;
\addplot [black]
table {%
1 0.158243477344513
1 0.143440246582031
};
\addplot [black]
table {%
1 0.18207985162735
1 0.217673540115356
};
\addplot [black]
table {%
0.8 0.143440246582031
1.2 0.143440246582031
};
\addplot [black]
table {%
0.8 0.217673540115356
1.2 0.217673540115356
};
\path [draw=black, fill=limegreen]
(axis cs:1.6,0.162960052490234)
--(axis cs:2.4,0.162960052490234)
--(axis cs:2.4,0.189069986343384)
--(axis cs:1.6,0.189069986343384)
--(axis cs:1.6,0.162960052490234)
--cycle;
\addplot [black]
table {%
2 0.162960052490234
2 0.148726224899292
};
\addplot [black]
table {%
2 0.189069986343384
2 0.228176116943359
};
\addplot [black]
table {%
1.8 0.148726224899292
2.2 0.148726224899292
};
\addplot [black]
table {%
1.8 0.228176116943359
2.2 0.228176116943359
};
\path [draw=black, fill=forestgreen]
(axis cs:2.6,0.176663398742676)
--(axis cs:3.4,0.176663398742676)
--(axis cs:3.4,0.209083318710327)
--(axis cs:2.6,0.209083318710327)
--(axis cs:2.6,0.176663398742676)
--cycle;
\addplot [black]
table {%
3 0.176663398742676
3 0.159581422805786
};
\addplot [black]
table {%
3 0.209083318710327
3 0.257616281509399
};
\addplot [black]
table {%
2.8 0.159581422805786
3.2 0.159581422805786
};
\addplot [black]
table {%
2.8 0.257616281509399
3.2 0.257616281509399
};
\path [draw=black, fill=darkgreen]
(axis cs:3.6,0.259650528430939)
--(axis cs:4.4,0.259650528430939)
--(axis cs:4.4,0.319266438484192)
--(axis cs:3.6,0.319266438484192)
--(axis cs:3.6,0.259650528430939)
--cycle;
\addplot [black]
table {%
4 0.259650528430939
4 0.231648206710815
};
\addplot [black]
table {%
4 0.319266438484192
4 0.407285451889038
};
\addplot [black]
table {%
3.8 0.231648206710815
4.2 0.231648206710815
};
\addplot [black]
table {%
3.8 0.407285451889038
4.2 0.407285451889038
};
\addplot [line width=1pt, black, dash pattern=on 9.25pt off 4pt]
table {%
-0.4 0.164279341697693
0.4 0.164279341697693
};
\addplot [line width=1pt, black, dash pattern=on 9.25pt off 4pt]
table {%
0.6 0.165372610092163
1.4 0.165372610092163
};
\addplot [line width=1pt, black, dash pattern=on 9.25pt off 4pt]
table {%
1.6 0.169936895370483
2.4 0.169936895370483
};
\addplot [line width=1pt, black, dash pattern=on 9.25pt off 4pt]
table {%
2.6 0.183742046356201
3.4 0.183742046356201
};
\addplot [line width=1pt, black, dash pattern=on 9.25pt off 4pt]
table {%
3.6 0.271404504776001
4.4 0.271404504776001
};
\end{axis}

\end{tikzpicture}